# An Agent-based Manufacturing Management System for Production and Logistics within Cross-Company Regional and National Production Networks

**Heinrich, S.**; **Durr, H.; Hanel, T. & Lassig, J.**
Department of Manufacturing Systems Engineering,
Chemnitz University of Technology, Germany
heinr@hrz.tu-chemnitz.de

*Abstract: The goal is the development of a simultaneous, dynamic, technological as well as logistical real-time planning and an organizational control of the production by the production units themselves, working in the production network under the use of Multi-Agent-Technology. The design of the multi-agent-based manufacturing management system, the models of the single agents, algorithms for the agent-based, decentralized dispatching of orders, strategies and data management concepts as well as their integration into the SCM, basing on the solution described, will be explained in the following.*

## 1. The Problem

One of the main goals of companies in the production industry is the satisfaction of demand on the market through the production and purchase of components under the use of own resources and if necessary through the establishment of close or loose co-operations with own company sectors or other companies:

Material, semi-manufactured products, single and standard parts, components etc. are produced by the company itself or are purchased; the operations which are necessary for it have to be provided with resources of manufacturing capacities (manufacturing systems, assembly units, devices etc.) as well as personnel capacities.

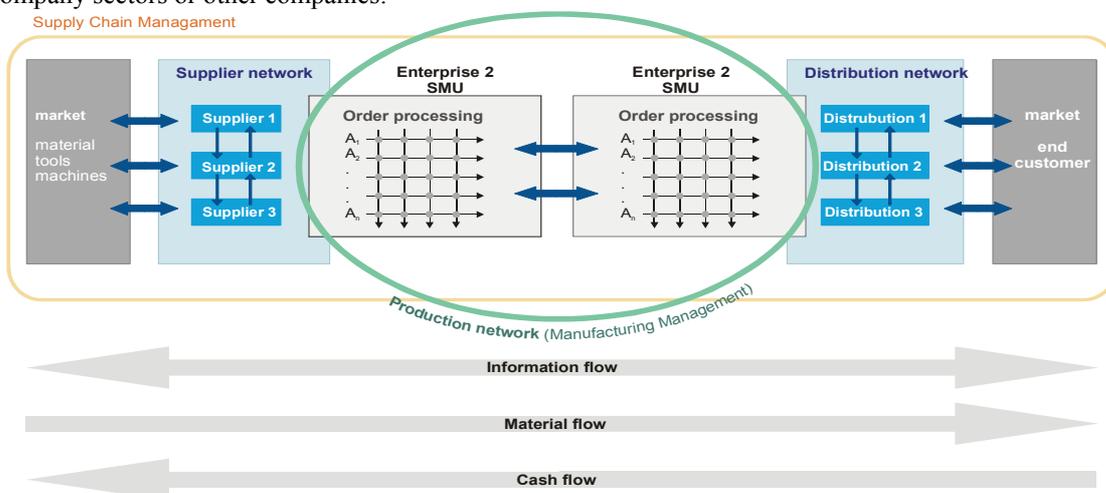

Fig.1. Integration of a Manufacturing Management System into a Supply-Chain-Management System for SME (Small and Medium-Sized Enterprises)

The concurrent incidental planning such as sourcing of material, scheduling, calculation of resource suitability and availability, valuation of resources according to microeconomic criteria, capacity determination as well as sequencing planning of job processing, and consideration of several manufacturing resources, can be modeled as allocation problems, where supplies are assigned respectively to a particular number of demand, so that existing restrictions can be observed.

Multi-agent systems (MAS) are a local method for effective solution of these problems, often characterized by additional conditions, which have to be considered. These problems are generally superior to mathematical optimizing algorithms, heuristics, Simulated Annealing or other methods and are in case of real world-scheduling problems difficult to solute and too time-consuming in calculation.

Illustration 1 shows the classification and possible functions of the manufacturing management systems in the company.

**2. The solution**

The developed solution model of the problem implies the agent tools with the following functionalities, described in chapter 3 more in detail:

- Agent-based model and algorithm bank for the generation of dispatching variants as well as an optimised allocation and processing of job orders and the generation of appropriate measures for the compensation of disturbances in real-time for companies or several company sectors connected through production networks; for applications within companies of the metal-cutting manufacture of components the algorithm for the adjustment of the machine parameters (cutting speed, infeed) described in (1) and (2) is to be implemented to the concrete manufacturing situation

- Optimized, event-related synchronization of essential local information flows (information transfer between the manufacturing segments of individual enterprises), as well as the synchronization of the global information flows in the production network (e.g. to suppliers, between subassembly and final assembly areas, to logistic service providers and other spatially separated company sectors) by supplier and data integration.

- Simultaneous flow chart planning within job order and disturbance management for all job orders which are currently in the production network, along with real-time assignment and state simulation of all resources.

- Schedule optimization under the use of different mathematical approaches

- Close-to-process recognition and management of deadline backlogs and capacity constraints as well as visualization of the production progress and cost development within the bounds of financial controlling.

- Innovative models and algorithms for resource and logistics substitution for the compensation of disturbances, such as problems with machines and systems, back orders, and the dispatching of additional, short-term orders in the running production.

**3. Design of an Agent-based Manufacturing Management System**

*3.1. Model-Theoretic Aspects*

Basically, there are two agent models representing the basic concept of this type of systems. As in the problem itself, the consumers are modeled as consumer agents on a software technical basis representing the job orders in a manufacturing environment; the supply of resources for the satisfaction of this demand is represented by supply agents. Thereby, a job agent represents a job order and a machine agent a production resource.

Each supply agent keeps a schedule, that resources cannot be allocated several times. A consumer agent has information about a complete job order and the operations. The solving strategy is the allocation of operations to be carried out to an appropriate supply agent through communication of the agents. The decision about which agent is to be considered for this communication is made by means of a specification vector defined for each agent.

In this specification vector all parameters necessary for the choice of resources, such as possible manufacturing or assembly processes, size of workshop, number of NC-axes, accuracy classes, current allocation situation etc. are taken into account.

At the moment design patterns are being worked out, which are aimed at making an efficient solution on a very high resource and order quantity possible. One important aspect in this context, is the establishment of working groups of several agents on several hierarchy levels. This is to continue to guarantee the local solution of standard problems in the future, on the one hand, and to make a solution for job orders which are difficult to dispatch and for which all resources available have to be taken into account almost world-wide possible. An experimental bottom-up-concept of certain parts of the agent algorithms, which cannot be specified more exactly at the moment, will follow as an essential part of these works.

The main advantages of the solution for manufacturing planning on the basis of MAS, outlined in the following, are:

- decentralization of the calculation of schedule, which is generally very time-consuming and

- a high flexibility concerning the reaction on unforeseen incidents in the production process, such as e.g. logistics problems, short-term dispatching of rush orders, tool damage or failure to comply with quality norms and changes in schedule connected with it, but also a high flexibility in respect to the scalability of these systems subject to the availability of production resources and the current job order revenue.

In the following, a design model of an agent-based manufacturing management system, the single agents and their interactions, its integration into the complete system as well as the methods for schedule optimization will be described more in detail.

*3.2. General Structure of the MAS*

The principal structure of an agent-based manufacturing management system with the main components, which are explained in chapter 3.3, is described in illustration 2 for several organizational levels.

During the implementation of the system, all resources of the company are reproduced as software agents and all necessary data is provided in a data basis. Additionally, the system administrator builds a software model of the organizational structure of the company i.e. the single manufacturing resources are allocated to certain sectors, which are assigned to organizational levels. Thus, as it is shown in illustration 2, for example, area 2a of organizational level 2 is able to access the machine agents $MA_1$ to $MA_j$, meanwhile area 3b of organizational level 3 only has the resources of the machine agents $MA_{i+1}$ to $MA_j$.

With the later definition of the job order agent, an allocation of the job order is made at first by the manufacturing management agents to an area corresponding with the generated specification vector resources available, is not possible, the request can be transmitted to the machine agents of different organizational structures (different company sectors, cooperating companies etc.). In that case, the sphere of influence of the relevant job order agents will be extended to organizational level 1. With that, it has access to the resources of the machine agents $MA_1$ to $MA_n$.

Possible alternative areas can be limited additionally by system restrictions on the basis of particular criteria (manufacturing resources available, current resource utilization etc.). In case of a simple milling order, for example, it is thus possible to exclude areas from different manufacturing processes at the outset, which increases the efficiency of the system.

The design of the multi-agent-system is flexible in double respects. With the integration of new manufacturing resources (new machines, facilities, technologies, assembly units etc.) or with the replacement of available manufacturing resources in different areas, the present structure can be adapted flexibly with the introduction of new or editing of available machine agents.

Moreover, the MAS can be simply adapted to changing organizational structures, which are, for example, conditional on the restruction of production areas of the company.

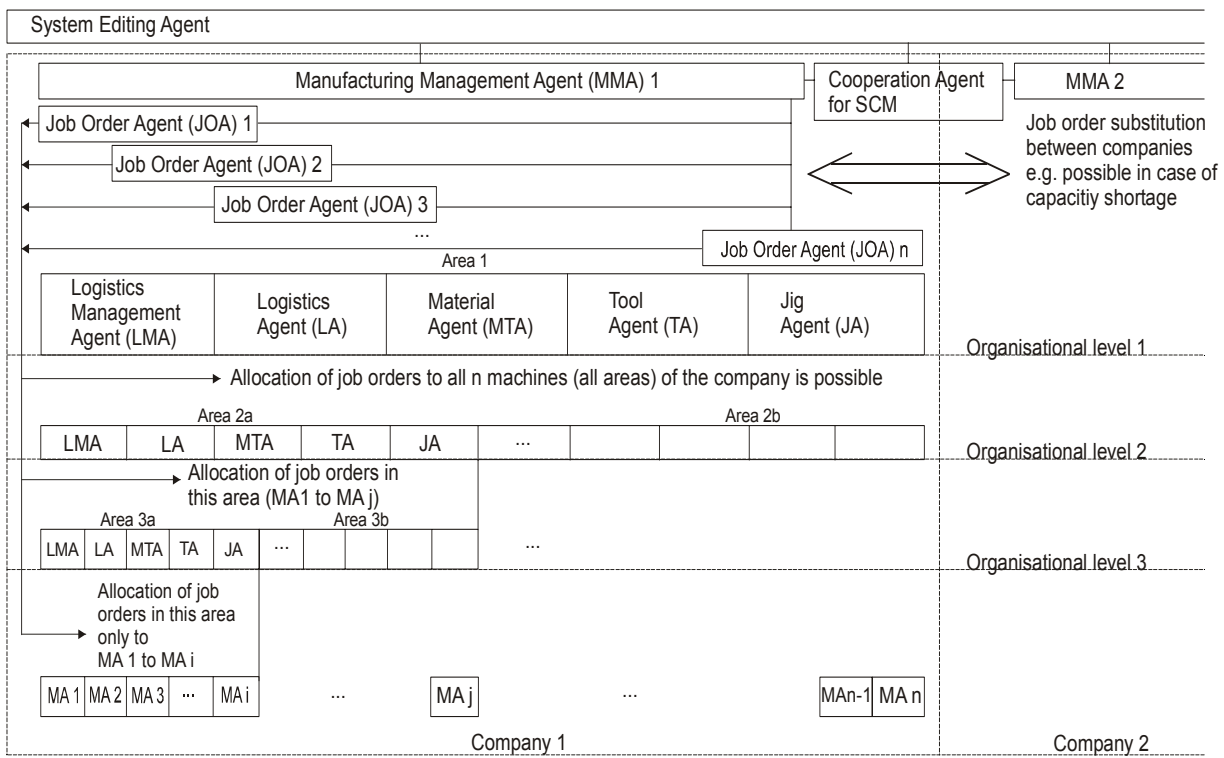

Fig. 2. General structure of an agent-based Manufacturing System with several organizational levels

(e.g. choice of areas according to manufacturing processes). If a processing of orders in due-date time within the specified areas, e.g. by full utilization of

Conditional upon the variability of the hierarchical levels of the MAS, it is variably applicable for companies of different size and is almost unlimited expandable.

This is an essential advantage, e.g. concerning the integration of several companies or company sectors into network organizations (e.g. regional or national production networks), which can be realized with a relatively low effort.
In order to guarantee the functionality of the complete system desired, the definition and implementation of the single agents, described in the following, is necessary.

*3.3. Single Agents and their Interactions*

*3.3.1 System Editing Agent (SEA)*
The SEA, which exists only once in the complete system, is responsible for the definition and editing of the complete system by the system administrator. The SEA writes the appropriate logical system structure in the data basis serving for the administration of restrictions and access rights for users and as reference basis for the single agents.
Consequently, by means of the SEA, new agents can be built, edited or deleted as it is e.g. with the purchase of new machines for a production area. Moreover, changes in the company structure, e.g. the allocation of machines and facilities to particular production areas, can be taken into account.

*3.3.2 Manufacturing Management Agent (MMA)*
For each new order the corresponding job order agents (JOA) are generated by the MMA. All necessary data is provided by the user or recalled from the data bank.
The order will be divided into several operations and includes the necessary process specifications (e.g. milling, drilling, assembly operations, etc.). Additionally, it is possible that process alternatives for the resource-dependent technology substitution are taken into account. If necessary, with the help of the MMA, requests to the particular JOA concerning the current state of process, the allocation to the manufacturing system etc. can be initiated and the answers can then be visualized according to the criteria specified by the user (machine scheduling, state of process etc.).

*3.3.3 Job Order Agent (JOA)*
For each new job order which is to be dispatched a JOA is generated (see illustration 3). This agent communicates with the production area to which it had been allocated by the MMA and the MA.
By means of significant parameters (processes, operations, number of axes, geometric variables etc.) appropriate MAs are determined with the aim to evaluate the feasibility of the job order at the single MAs.
Additionally, the JOA edits requests of the MMA in regard to scheduling, job order progress, machine scheduling of particular job orders and balances the transfer of the single machines with the logistic agent (LA).

*3.3.4 Logistic Management Agent (LMA)*
Each LMA is allocated to one production area and records the logistic parameters (transport capacities and transport times) of the production area allocated into the data basis.

*3.3.5 Logistic Agent (LA)*
One LA is assigned to each production area. The LA edits requests of job order agents (JOA), tool agents (TA), material agents (MA), jig agents (JA) with the aim to evaluate the feasibility.

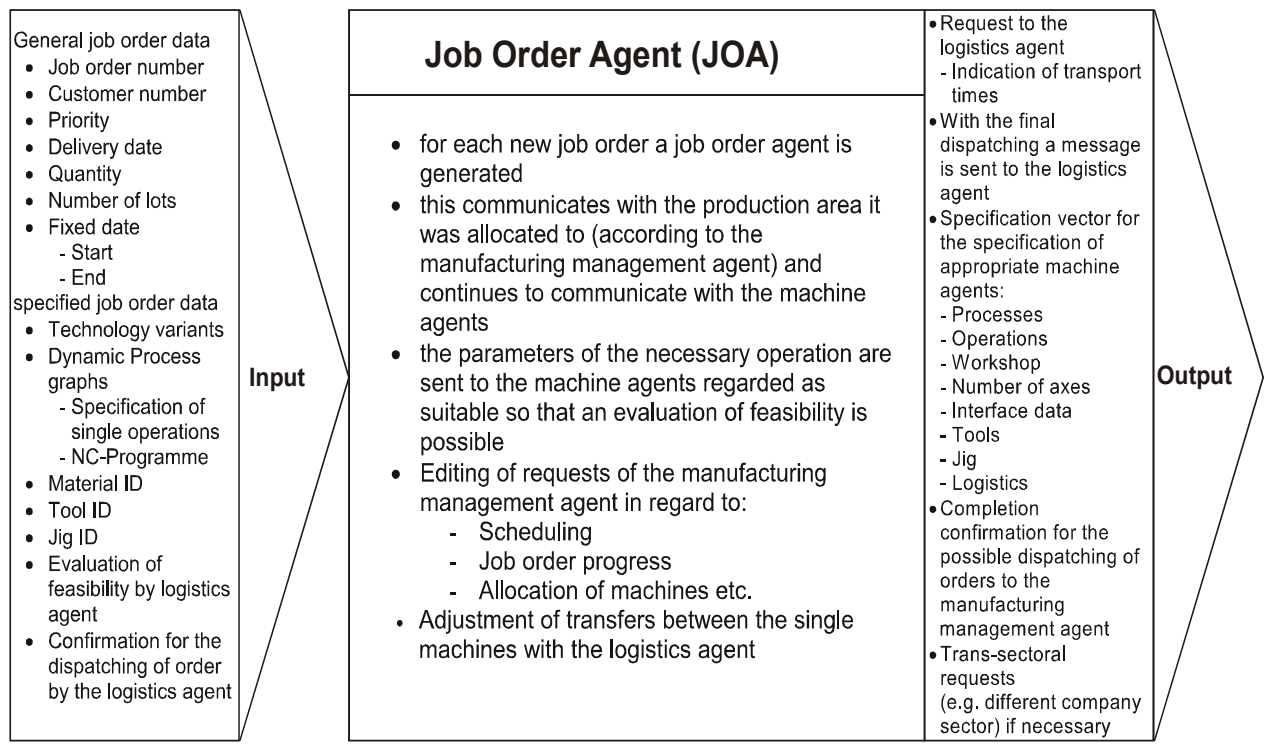

Fig. 3. Functionality and data management of the job order agent

All necessary data is provided by the data basis of the LMA. For all actually dispatched job orders, a logistic schedule is kept, which, if necessary, can be visualized the user according to particular criteria.

*3.3.6 Tool Agent (TA)*
Each TA is allocated to one production area. The TA accepts the job order specific tool requirements of the MA, examines if they can be implemented and makes the tools after dispatching of orders in the course of the preparation of orders available.
During order processing, the tooling insert is monitored and in case of disturbances (e.g. tool damage, end of tool life), a message is sent to the respective MA.
The TA keeps the minutes about tools that are still available (in stock) and about those which have been already taken into account for order processing.

*3.3.7 Jig Agent (JA) and Material Agent (MTA)*
These agents have the same configuration and functions as the Tool Agent but in regard to jigs and devices and material.

*3.3.8 Machine Agent (MA)*
One MA is allocated to each machine. The MA is the negotiating partner for the JOA. It keeps the machine scheduler for the single processing operations allocated to the machine and is responsible for the examination of and guarantee for the feasibility of the predetermined operations during dispatching of orders under the use of its supply agents.
Additionally, a data exchange with the TA, the JA and MTA is necessary.
In case of disturbances, the MA sends a message to the JOA.

**4. Dispatching Strategies for Agent-based Dispatching of Orders**

*4.1. General Aspects*
The dispatching of orders comprises two aspects, which are principally contrary to each other. On the one hand, it aims at short response times of the system if a new order is to be dispatched; on the other hand, it is, due to a high complexity, theoretically impossible to determine the implementation of an order by means of complex calculation in this short period of time.
Thus, it appears to be useful, to provide several dispatching strategies which use certain assessment indexes as a basis. This has the advantage, that short response times can be guaranteed through fast calculation.
If a dispatching of orders under these conditions is not possible, a second dispatching attempt with a different strategy can be made. Within this work, those strategies are to be provided which put the planning process to the fore of the consideration and disregard optimization aspects (e.g. use of the same configuration of the preceding operation) more and more:
1. Dispatching Strategy „OPT"
2. Dispatching Strategy „Force"
3. Dispatching Strategy „X-Competition"
4. Dispatching Strategy „Wait-X"
5. Dispatching Strategy „Manual".

*4.2. Assessment Indexes*
For the selection of the most appropriate machine agents, several **assessment indexes** are used, depending on the particular dispatching strategy. These indexes can be calculated by the machine agent and give the job agent both information about the machine qualification and the quality of the calculated dispatching time concerning several target criteria for the operation (e.g. robustness or minimal processing time).

*4.2.1. Machine index*
The machine index evaluates, to what extent the machine over-fulfills the requirements of the operation. For it, all parameters are to be evaluated which can vary in dimension with the different machines of a division (e.g. number of axes). Parameters which have exclusively the status "available" and "unavailable" (e.g. DNC Net) will not be taken into account.

*4.2.2. Robustness index*
During order definition, a **robustness time** is determined for each operation, which, if possible, is to be dispatched to the absolute operation time.
The robustness time provides a kind of excess charge, which is aimed at compensating short term disturbances or deadline shifts of the operation. The robustness index evaluates, to what extent the robustness time of a determined time gap can be implemented for an operation.

*4.2.3. Position index*
If there is an appropriate time gap determined, the position of the operation (inclusive robustness time) can be evaluated within this time gap. For it, the position index evaluates the remaining time from the end of the operation to the end of the time gap; if it is, regarding the average, too big or to small for further orders. The specific **average processing time** (APT) of the machine is used as a reference for a still "useable" time gap.
APT can be determined, for example, with the help of statistical enquiries and complies flexibly with the orders which have been carried out previously on the machine.

*4.2.4. Setting-up index*
The setting-up index evaluates a determined time gap in regard to the degree of probability with which the setting-up of the machines of the previous operation can be used again for the operation which has to be dispatched. APT can be considered to be a reference time again, since with the increasing interval of the operations, the probability that a further operation can be dispatched in this time gap later on increases and, therefore, the reuse of the setting up is impeded.

### 4.2.5. Time slot index

Depending on the direction of the dispatching (forwards and backwards), the time slot index determines the position of the operation within the time slot, which is set by the order agent and, thus, affects the processing time of the order.

### 4.2.6. Total index

For each determined operation time, a total index is calculated out of all available indexes (machine index, robustness index, position index, setting-up index and time slot index). During that process, the dispatching strategy, which has been determined during the order release, decides, which of the five indexes have to be considered in the total index.

## 4.3. Dispatching Strategies

### 4.3.1. Dispatching strategy „OPT"

During dispatching strategy OPT, the dispatching of orders is made backwards, in other words, it is orientated towards the latest finish time of an order (complies normally with the desired date of the customer).
The advantage of the use of the backwards dispatching strategy is its ability to achieve shortest processing times. During dispatching strategy OPT, the orders and the operations linked to them are attempted to be dispatched in the schedule under consideration of optimisation aspects, i.e. according to the strategy, the machine agent allocates each of the five assessment indexes to one total index during the determination of the most appropriate dispatching date.
The weighting which is chosen in this process is variable and flexibly adaptable to the target criteria of the company. Thus, the time slot index, for example, can flow in the total index very high weighted if the major goal is the implementation of short processing times, or the robustness index in an analogous manner if the main attention is turned to a stable and robust machine schedule.

### 4.3.2. Dispatching strategy „Force"

Contrary to dispatching strategy OPT, the termination in dispatching strategy "Force" is forwards since, here, the optimal dispatching is not the focus of attention but rather the achievement of a dispatching process itself.
In "Force" the weighting of indexes, which are provided by the machine agents, differs from that of OPT. The machine and position indexes are not taken into account by the machine agents, but rather the dispatching time will be chosen from the available possibilities which offer the best total index of robustness and time slot index.
An additional option, which the strategy provides for the manufacturing management, is the deactivation of the robustness index. With this option, chances of a successful dispatching of operations can be consistently increased, linked to the disadvantage of a lower robustness of the schedule.

### 4.3.3. Dispatching strategy „X-Competition"

In X-Competition, the competition situation between different orders is used as dispatching criterion. The "X" stands for a priority input, which has to be chosen by the manufacturing management if this dispatching strategy is used.
In principle, the manufacturing management assigns priorities between 1 and 5 to the single orders, whereby priority 5 is defined as the highest category.
During the dispatching of orders in the MAS, orders of low priority, marked as "X", are ignored and their dispatching times are considered as free time gaps. However, within the scope of dispatching strategy "X-Competition", the operation specific job priority is only accounted for as the basic factor of a *dynamically calculated job priority*.

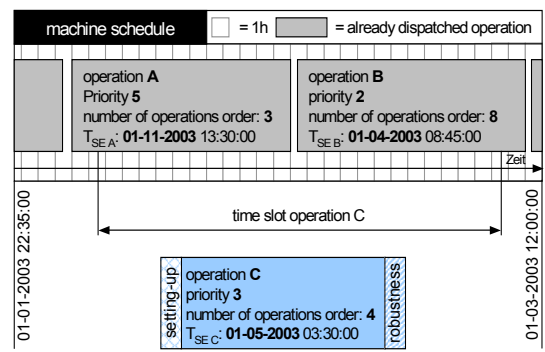

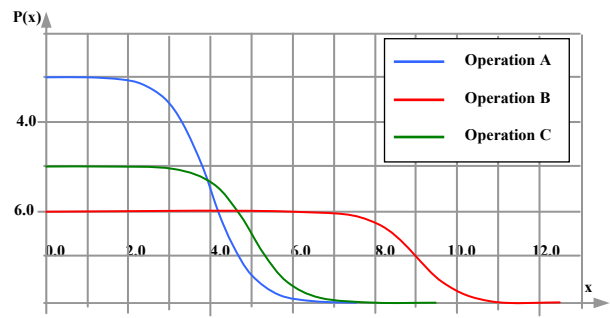

Fig. 4. Dispatching strategy X-Competition and dynamic order priority

### 4.3.4. Further dispatching options for „Force" and „X-Competition"

In both, dispatching strategy Force and X-Competition, the actual dispatching of orders has priority over optimization criteria, such as the aspect whether the setting-up of the preceding operation can be used for the machine a second time. For this purpose, three operations are introduced, which make it possible to dispatch an order even if there is a sufficient, coherent time gap in the schedule of the machine agent lacking.

- **Overdraft** makes a dispatching of orders after the end of a shift in a company possible. For example, if there is an order to be dispatched in a time gap before the end of a shift, and the order exceeds it for less than 30 minutes (this time is as a parameter variable depending on the specific requirements of the company), the order will be dispatched even with priority 5 and, at priority 4, the manufacturing management is asked if the dispatching is to be implemented. For orders with priority 3-1 this option is not available.

- **Shift splitting** makes it possible that an order can be dispatched after a shutdown of the machine, i.e. the order is stopped at the end of the shift and can be continued at the start of the next shift on the same machine. As a suboption the following ways of interrupting the process are possible:
  a) in the lot or transport lot
  b) after a full transport lot or
  c) after a certain transport lot number

- **Long-time-Splitting** refers to the so-called „Langläufer" (long-running operations) and, thus, to operations which are difficult to dispatch during a high capacity utilization of the machines. If there is an operation which has a total processing time exceeding x minutes (e.g. 1200 min or 20h), the operation can be split. The operation can automatically be split up into shorter operation parts which have to be longer than y minutes (e.g. 300min), and the partial lots can be dispatched on different machines and the time intervals between these partial lots can also be bigger.

*4.3.5. Dispatching strategy „Wait-X"*

The basic idea of "Wait-X" is the holding up of an order in the system with the condition that it has to be dispatched by a fixed deadline "X" (e.g. if through rescheduling of other orders or optimization steps time gaps in the schedules of the machines become available which correspond to the time necessary for the order to be dispatched).

In case, the dispatching of an order cannot take place within the fixed time frame, the manufacturing management gets a message after the time limit has been reached.

*4.3.6. Dispatching strategy „Manual"*

An order which cannot be dispatched by means of the available strategies accumulates in the manufacturing management. The dispatching strategy "Manual" follows the direct interaction with the user, i.e. the technologist or work scheduler tries to dispatch the order manually with the help of the system. For this process several functions are available to the user:

- **Explicit Splitting** → here, the manufacturing management determines in how many different orders the original order has to be split.
- **Manual Splitting** → here, the order is to be split in the way that according to the manufacturing management appropriate time gaps can be used (splitting according to available schedule). Additionally, the manual modus allows for:

a) **Change of Restrictions** (e.g. priority or completion date) with a repeating planning process afterwards.
b) **Deletion** of an order which had already been dispatched and replacing by the desired order.
c) **Outsourcing**, i.e. passing of the order forward to the SCM-agent with the goal of dispatching the order in another company or company sector

During the process of the last-mentioned option "Outsourcing", one part of an order or the complete order is passed forward to a company of the Supply Chain. The coordination of the cross-company dispatching of orders is managed by the Supply-Chain-Management agent.

## 5. Optimization strategies

The optimization strategies are part of the combinational optimization. A modified branch & bound algorithm (see (4) and (5)) is taken as a basis, whose aim it is, to determine a preferably short schedule.

The primary schedule is determined by the already dispatched operations. The principle of the algorithm is the freezing of certain operations in the schedule and the rescheduling of the remaining operations under consideration of existing dependencies (material, tools etc.).

During the process of rescheduling one of the in paragraph 4 described dispatching strategies is used.
The optimization task is:

*find*:
make-span := {time(t) + d(t)} → MIN!

*with*:
$\forall\ t_1, t_2 \in T$: precede $(t_1, t_2) \rightarrow time(t_2) \geq time(t_1) + d(t_1)$

$\forall\ t_1, t_2 \in T$: use$(t_1)$ = use$(t_2) \rightarrow$ [time$(t_2) \geq$ time$(t_1)$+ d(t1)] $\vee$ [time$(t_1) \geq$ time$(t_2)$+d$(t_2)$]

*with*:
| | |
|---|---|
| time(t) | starting time of task t |
| d(t) ≥ 0 | duration of task t |
| T | number of tasks |
| use(T) | amount of resources to be used |
| precede($t_1, t_2$) | $t_1$ must be completed before $t_2$ can proceed |

The Make-span is the maximum duration of the running schedule and complies with the duration from the starting point of the earliest dispatched operation till the finishing point of the last dispatched operation. The optimization is carried out in three steps:

*Level 1: Repair Step and Swaps*
Swapping of operation pairs, meaning that operations pairs which would show a shorter schedule after the swapping process, will be swapped. The swapped operations will be frozen in the schedule, the remaining ones will be dispatched again.

*Level 2: Basic Shuffle*

The operations of the machine which show the biggest time gap between two operations will be frozen, the remaining ones will be dispatched again. This proceeding will be carried out on all machines.

*Level 3: Vertical Shuffle*
All operations within a random determined intervals will be frozen, when they all are part of an order.

*Level 4: Horizontal Shuffle*
Out of the machines which do not cause problems all operations of the two machines with the biggest time gaps will be frozen in.

With this optimization method an average schedule optimization of approximately 10 to 35 percent has been reached in a number of test runs depending on the specific usage conditions and dispatching criteria.

Due to the relative time intensity of the optimization it is started automatically in "neutral phases" of the MAS, meaning during phases in which no active dispatching is taking place or in phases off the operation time.

The manufacturing management is informed via optimised schedules and it is possible to deny optimisations and to restore the schedule to its initial state.

## 6. Conclusion and Prospects

Saturated markets, the difficulty to make sales figures predictable as well as highest adherence to delivery times and production quality are only a few of the features of the competition, companies are being exposed to. Therefore, manufactures are forced to mutability as well as to the development of dynamic production and organizational structures, which can be realized in its entirety with the solution presented.

The main advantages of the solution presented are the decentralization of the scheduling, the high flexibility concerning the reaction to all situations occurring in the production and the high flexibility in the scalability of the system depending on the concrete situation of the company. Thus a use in companies of the manufacture of components and the chip industry as well as in logistics fields is possible.

A significant prospective opens up with a process and resource orientated dispatching and rescheduling of new job orders via all necessary production units of one or more companies connected each other through network linked enterprises as well as the dynamic optimization of schedules under consideration of the relevant order-conditional relations.

Thus, it is possible to lay the foundation for a more efficient utilization of resources available in the company network already in the dispatching phase.

New opportunities open up with the use of a Disturbance-Detection-Agent, which still has to be developed and which is, with the help of data-mining methods on the basis of in real-time assigned process parameters, able to recognize disturbances already very early, to generate warning messages effectively and to pass them to the manufacturing management.